\documentclass[sigconf]{acmart}

\renewcommand\footnotetextcopyrightpermission[1]{} 
\acmConference{}
\acmYear{}
\acmDOI{}
\acmISBN{}
\acmBooktitle{}

\AtBeginDocument{%
  }

\usepackage{geometry}
\usepackage{tabularx}
\usepackage{booktabs}
\usepackage{array}
\usepackage{multirow}
\usepackage{caption}
\usepackage{pdflscape}
\usepackage{pgfplots}
\usepgfplotslibrary{groupplots}
\usepackage{afterpage}
\usepackage{arydshln}
\usepackage{graphicx}
\usepackage{datatool}
\usepackage{tikz}
\pgfplotsset{compat=1.17}
\usepackage{subcaption}
\usetikzlibrary{patterns}
\usepackage{graphicx}
\usepackage[title]{appendix}
\pdfoutput=1
\begin{document}

\title{Enhancing Financial Question Answering with a Multi-Agent Reflection Framework}

\author{Sorouralsadat Fatemi}
\email{sfatem6@uic.edu}
\affiliation{%
  \institution{University of Illinois at Chicago}
  \city{Chicago}
  \state{IL}
  \country{USA}
}
\author{Yuheng Hu}
\email{yuhenghu@uic.edu}
\affiliation{%
  \institution{University of Illinois at Chicago}
  \city{Chicago}
  \state{IL}
  \country{USA}
}









\begin{abstract}
While Large Language Models (LLMs) have shown impressive capabilities in numerous Natural Language Processing (NLP) tasks, they still struggle with financial question answering (QA), particularly when numerical reasoning is required. Recently, LLM-based multi-agent frameworks have demonstrated remarkable effectiveness in multi-step reasoning, which is crucial for financial QA tasks as it involves extracting relevant information from tables and text and then performing numerical reasoning on the extracted data to infer answers.
In this study, we propose a multi-agent framework incorporating a critic agent that reflects on the reasoning steps and final answers for each question. Additionally, we enhance our system by adding multiple critic agents, each focusing on a specific aspect of the answer. Our results indicate that this framework significantly improves performance compared to single-agent reasoning, with an average performance increase of 15\% for the LLaMA3-8B model and 5\% for the LLaMA3-70B model.
Furthermore, our framework performs on par with, and in some cases surpasses, larger single-agent LLMs such as LLaMA3.1-405B and GPT-4o-mini, though it falls slightly short compared to Claude-3.5 Sonnet. Overall, our framework presents an effective solution to enhance open-source LLMs for financial QA tasks, offering a cost-effective alternative to larger models like Claude-3.5 Sonnet.
\end{abstract}



\maketitle

\section{Introduction}
The analysis of financial documents, such as SEC filings, is crucial for assessing business and company performance. Analyzing these financial documents requires advanced expertise in financial knowledge, the ability to reason across both tabular and textual data sources, and the capability to perform complex numerical reasoning \cite{zhu2021tat}. Recent studies have explored the effectiveness of various models and approaches in comprehending financial documents through Question-Answering (QA) tasks \cite{chen2021finqa, chen2022convfinqa}.
Financial QA datasets typically contain hybrid data, comprising both structured tabular information and unstructured textual content. This heterogeneous nature of financial documents presents unique challenges for natural language processing systems. Effectively comprehending and answering questions in such hybrid contexts requires not only an understanding of the intricate relationships between tabular data and narrative paragraphs but also the ability to perform advanced numerical reasoning. These skills are essential for tasks such as interpreting financial statements, analyzing performance metrics, and drawing insights from complex financial reports \cite{chen2021finqa, chen2022convfinqa, zhu2021tat}. The complexity of financial QA extends beyond basic data comprehension. To effectively answer each question, relevant numerical data must be accurately extracted from both textual descriptions and tabular representations. Furthermore, deriving answers often involves combining and applying various mathematical operations—including counting, sorting, comparing, and basic arithmetic (addition, subtraction, multiplication, and division) \cite{zhou2022unirpg}. This multi-step process, involving complex data interpretation and manipulation, renders financial QA tasks particularly challenging, requiring advanced numerical reasoning capabilities that go beyond traditional natural language processing techniques.


To address these challenges in financial QA tasks, researchers have proposed various techniques. Some studies have employed sequence tagging to extract relevant numbers from tables and pertinent spans from text, facilitating semantic inference \cite{zhu2021tat}. Another study proposed a hierarchical approach using multi-level graphs to model semantic relationships between quantities, dates, and text. This method aims to improve the extraction of relevant information from hybrid contexts \cite{zhu2023soargraph}. Furthermore, some researchers developed heterogeneous graphs to represent the relationships between different data types, capturing the correlation between tables and texts and aggregating them efficiently \cite{lei2022answering}. While these approaches have shown promising results, their effectiveness is often constrained by the considerable complexity involved in data preprocessing and graph construction to encode numbers and texts from various data types. Moreover, the performance of these approaches is often limited by the capabilities of the underlying language models typically employed in these studies, such as encoder-only or encoder-decoder architectures.

Recently, Large Language Models (LLMs) have demonstrated remarkable success in various reasoning tasks, including mathematical and commonsense reasoning \cite{zhao2023survey}. Despite these advances, LLMs still face challenges with complex reasoning tasks \cite{xu2023large, zhu2022solving, gou2023critic}. To address these challenges, researchers have begun developing LLM-based multi-agent frameworks capable of executing intricate multi-step decision-making and reasoning tasks \cite{yao2022react, hao2023reasoning, richards2023auto}. These frameworks have attracted considerable attention for their potential to enhance mathematical and strategic reasoning across diverse applications, such as sequential decision-making, evidence-based QA, and language reasoning \cite{qin2023toolllm, du2023improving, zong2024triad}. Notably, some of these frameworks incorporate reflective agents that employ an iterative refinement process. In this approach, the LLM agent improves its answers based on previous outputs and feedback, enabling more sophisticated reasoning for complex tasks \cite{xing2024designing, lin2024interpreting, renze2024self, shinn2024reflexion}. 

However, to the best of our knowledge, the effectiveness of these multi-agent systems has not been explored in the context of financial QA tasks. Financial QA is an integrated task requiring arithmetic and financial knowledge, unlike other reasoning tasks that focus on a single specific ability, and thus necessitates multi-step reasoning. This gap presents an opportunity to investigate how these LLM-based agents can be applied to financial QA tasks.


To address this research gap, we propose a multi-agent framework for financial QA tasks. Within this framework, we implement agents built upon a large language model (LLM) core and examine three settings.

The framework begins with an expert agent that leverages chain-of-thought (CoT) prompting for data extraction from tables and text, followed by mathematical reasoning to generate answers. Building upon this foundation, the second setting adds a critic agent that analyzes the expert agent's responses to enhance reasoning in future attempts.
To further improve the refinement process, we divide the task into two sub-tasks, assigning them to separate critic agents. Their reflection feedback is then passed to the first agent for re-evaluation of the initial response.

\begin{figure}[h]
  \centering
  \includegraphics[width=6cm]{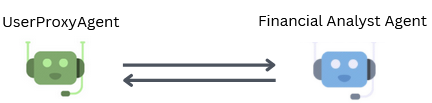}
  \caption{Single-Agent Setting}
  \label{fig1}
\end{figure}
\begin{figure}[h]
  \centering
  \includegraphics[width=\linewidth]{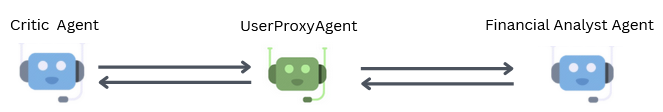}
  \caption{Two-Agent Setting}
  \label{fig2}
\end{figure}
\begin{figure}[h]
  \centering
  \includegraphics[width=\linewidth]{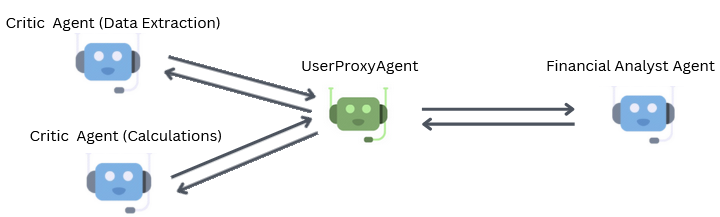}
  \caption{Three-Agent Setting}
  \label{fig3}
\end{figure} 

To demonstrate the advantages of our multi-agent system, we conducted experiments with various agent configurations using different LLMs, including LLaMA3-8b and LLaMA3-70b, on three popular finance domain benchmarks for hybrid tabular and textual QA: FinQA, ConFinQA, and TAT-QA.
Our experimental results indicate that our multi-agent framework with critic agents can enhance numerical reasoning capabilities compared to single-agent frameworks, consistently improving performance across the three datasets. Dividing refinement tasks between agents also improves performance, though by a smaller margin. Significantly, this approach achieves results comparable to state-of-the-art fine-tuned LLMs without the costs associated with dataset preparation and fine-tuning. Furthermore, it performs on par with proprietary larger models such as GPT-4o-mini and Claude-3.5-Sonnet, while circumventing their higher inference costs.

\section{Related Work}
Research on tabular and textual QA in the finance domain has primarily focused on decomposing tasks into multiple steps, generating intermediate results that guide the final answer \cite{zhou2022unirpg, lei2022answering, zhu2023soargraph}. Earlier studies, constrained by the limitations of encoder-only or encoder-decoder models in QA and reasoning tasks, developed novel methods to overcome these challenges.
One study proposed a semantic-parsing-based approach, substituting the prediction vocabulary with an operation set and integrating a copy mechanism into the BART model. This allowed for the retrieval of reasoning-related numbers or text from tables and documents when decoding programs. The researchers constructed programs based on annotated derivations for training instances \cite{zhou2022unirpg}. Another study introduced a relational graph modeling method to align questions, tables, and paragraphs, framing numerical QA over hybrid table-text content as an expression tree generation task \cite{lei2022answering}.
A more recent study modeled semantic relationships and dependencies among questions, table cells, text paragraphs, quantities, and dates using hierarchical graphs. At the lower level, graphs were built to model value magnitude, comparison information, and semantic relations in text. The higher level captured semantic relationships and dependencies \cite{zhu2023soargraph}. While effective, these methods require extensive data preparation, such as graph or tree structures, to extract semantic relationships from hybrid financial data.

The emergence of LLMs has demonstrated impressive language understanding and generation capabilities \cite{bubeck2023sparks, touvron2023llama, openai2022chatgpt, openai2023gpt4}. However, LLMs still struggle with multi-step and numerical reasoning tasks, encountering challenges such as hallucination and reasoning errors \cite{xu2023large}. To address these limitations, some studies have focused on further pre-training or fine-tuning smaller open-source LLMs.
In the finance domain, a recent study proposed a three-step approach (Extractor, Reasoner, and Executor) to enhance LLMs' multi-step inference abilities. The researchers constructed a step-wise pipeline dataset and fine-tuned various sizes of LLaMA models (7B, 13B, and 70B) \cite{zhu2021tat}. While promising, this approach faces challenges related to the high computational costs and memory requirements associated with fine-tuning LLMs.
More recently, inspired by the Society-of-Mind concept, multi-agent discussion frameworks such as Multi-Agent Debate (MAD) and ReConcile have emerged \cite{minsky1988society, du2023improving, cobbe2021training, chen2023reconcile}. These frameworks involve multiple agents powered by LLMs, engaging in discussions on given topics or tasks to improve reasoning abilities by emulating human discussion processes. This approach has shown promising results in various tasks, including problem-solving, evidence-based reasoning, and knowledge-based question-answering \cite{zong2024triad, schimanski2024towards}.

The efficacy of this multi-agent approach for financial QA tasks remains unexplored in current research. Our study investigates the potential of designing a multi-agent framework with critic agents built on smaller open-source LLMs to improve their numerical reasoning capabilities in financial contexts.

\begin{table*}
\centering
{\scriptsize\ttfamily
\begin{tabular}{|p{0.49\textwidth}|p{0.49\textwidth}|}
\hline
\multicolumn{1}{|c|}{\textbf{Financial Analyst Agent System Message}} & \multicolumn{1}{c|}{\textbf{Critic Agent System Message}} \\
\hline
You are a financial analysis agent specializing in interpreting earnings reports and financial statements. Your task is to answer specific financial questions based on the given context from financial reports.

When answering questions:
\begin{itemize}
\item Carefully read and analyze the provided financial information.
\item Extract the relevant data points needed to answer the question from the table or text provided.
\item Perform any necessary calculations.
\item Remember to be precise in your calculations and clear in your step-by-step explanation. Maintain a professional and objective tone in your response.
\item Use only the information provided in the context. Do not introduce external information.
\item Provide the answer in the unit specified in the question (million, percentage, or billion). If no unit is specified, use the most appropriate unit based on the context and question.
\end{itemize}
&
You are a reflective AI agent tasked with critically analyzing financial analyses. Your job is to review a given context, question, and the response provided by another agent. Then, you must reflect on the analysis and provide a detailed critique.

Your tasks are:
\begin{itemize}
\item Carefully read the provided context, question, and response.
\item Analyze whether the question was correctly understood and addressed.
\item Verify if the correct numbers were extracted from tables and text in the context. Double-check these numbers against the original context.
\item Check the accuracy of the calculations in each step provided. Recalculate each step to ensure correctness.
\item Verify if the logic of the steps provided is sound and appropriate for answering the question.
\item Assess if the final answer calculation is correct. Perform the calculation independently to confirm.
\end{itemize} \\
\hline
\end{tabular}
}
\caption{System Message for Financial Analyst Agent (Left), and System Message for Critic Agent (Right)}
\label{fig:system_messages}
\end{table*}

\begin{table*}
\centering
{\scriptsize\ttfamily
\begin{tabular}{|p{0.49\textwidth}|p{0.49\textwidth}|}
\hline
\multicolumn{1}{|c|}{\textbf{Critic Agent-Data Extraction System Message}} & \multicolumn{1}{c|}{\textbf{Critic Agent-Calculations System Message}} \\
\hline
You are a meticulous financial analyst and critic. Your task is to review the response provided by another agent regarding financial calculations and provide feedback on its accuracy and completeness. Pay close attention to the following aspects:

Review the given response for:
\begin{itemize}
\item Question Comprehension: Does the response correctly understand the original question?
\item Data Extraction: Are all relevant numbers accurately extracted from the provided text/tables?
\end{itemize}

Focus only on these two aspects. Do not evaluate calculations or provide additional analysis.
&
You are a meticulous financial analyst and critic. Your task is to review the response provided by another agent regarding financial calculations and provide feedback on its accuracy and completeness. Pay close attention to the following aspects:
\begin{itemize}
\item Calculation Steps: Confirm that all calculation steps are correct.
\item Calculation Accuracy: Verify the accuracy of all calculations, including intermediate and final results.
\item Unit Consistency: Ensure the final answer's unit matches what the question requires.
\end{itemize} \\
\hline
\end{tabular}
}
\caption{Critic Agent-Data Extraction System Message (Left), and Critic Agent-Calculations System Message (Right)}
\label{fig:critic_agent_messages}
\end{table*}

\section{Methodology}
Multi-agent communication enables autonomous dialogue between LLM-powered agents, each guided by specific prompts. Our research applies this paradigm to financial QA tasks, employing two specialized agents: a financial analyst expert and a critic. The expert interprets questions and analyzes hybrid financial data, while the critic evaluates and refines these responses. This iterative, collaborative approach enhances the accuracy of financial analysis, particularly for queries requiring advanced numerical reasoning. 
\subsection{Multi-Agent QA Framework}
Our multi-agent QA framework experiments with three different settings:

\textbf{Single-Agent Setting}: 
This framework, as illustrated in Figure \ref{fig1}, consists of two primary components: an executor agent and a financial expert agent. The financial expert agent is built upon a large language model and is equipped with specific financial analysis capabilities through a crafted system message, effectively providing the agent with a specialized persona. This system message is shown in Table \ref{fig:system_messages}. Despite having two components, we categorize this as a single-agent system because it utilizes only one LLM-based agent (the financial expert) guided by the specific system prompt.

The process can be formalized as follows:

\begin{equation}
A = \text{agent}_\text{expert}(T, D, Q, P_\text{expert})
\end{equation}
In this formulation, $A$ represents the output generated by the agent, which includes a detailed, step-by-step response outlining the agent's thinking process, the numbers extracted from the provided data, the mathematical operations performed, and the final answer to the question. The inputs to the agent include $T$, which represents the input text data; $D$, which denotes any tabular data provided; and $Q$, which is the specific question. $P_\text{expert}$ refers to the system prompt. The term $\text{agent}_\text{expert}$ represents the LLM-based financial expert agent, which handles these inputs to generate the final output $A$.

\textbf{Two-Agent Setting}:
This two-agent setting, illustrated in Figure \ref{fig2}, enhances our framework by incorporating a critic agent, guided by a system message shown in Figure \ref{fig:system_messages}, to refine the analysis process. The initial response from the financial expert agent is obtained in the single-agent setting. Subsequently, the critic agent, represented by $\text{agent}_\text{critic}$, receives the original question $Q$, input text $T$, tabular data $D$, its specific prompt $P_\text{critic}$, and the financial expert's initial answer $A$. The critic agent then generates a reflective or critical comment $R$, as formalized in Equation (2):
\begin{equation}
R = \text{agent}_\text{critic}(T, D, Q, P_\text{critic}, A)
\end{equation}
This comment $R$ is then fed back to the financial expert agent, $\text{agent}_\text{expert}$, along with its system prompt $P_\text{expert}$. The financial expert agent uses this feedback to refine its reasoning, calculations, and analysis, producing a revised answer, $A_\text{revised}$, as expressed in Equation (3):
\begin{equation}
A_\text{revised} = \text{agent}_\text{expert}(R, P_\text{expert})
\end{equation}
The detailed workflow of this two-agent system is illustrated in Figure \ref{fig7} in Appendix~\protect\ref{appendix}. As depicted in this figure, the fourth box demonstrates the critic agent's role in providing feedback on the steps outlined in the financial expert's initial answer. In this particular example, the critic agent identifies that the last step in the financial expert's reasoning is incorrect. Subsequently, upon receiving this feedback, the financial expert agent revises its answer, correcting the identified error and providing the accurate steps along with the final answer. This example underscores the effectiveness of the iterative refinement process, demonstrating how the interplay between the critic and financial expert agents leads to more accurate answers.
\textbf{Three-Agent Setting}:
In this configuration, as shown in Figure \ref{fig3}, we leverage two specialized critic agents, each focusing on refining specific aspects of the response. The refinement process is divided into two sub-tasks, with each critic agent specializing in one aspect. Both critic agents have their specific system messages, as depicted in Figure \ref{fig:critic_agent_messages}.

The process is formalized as follows:

\begin{equation}
R_1 = \text{agent}_\text{critic1}(T, D, Q, P_\text{critic1}, A)
\end{equation}

\begin{equation}
R_2 = \text{agent}_\text{critic2}(T, D, Q, P_\text{critic2}, A, R_1)
\end{equation}

\begin{equation}
A_\text{revised} = \text{agent}_\text{expert}(R_1, R_2, P_\text{expert})
\end{equation}

The first critic agent, represented by $\text{agent}_\text{critic1}$, is provided with the tabular data $D$, text data $T$, the question $Q$, and the initial answer $A$ from the financial expert agent. This agent is specifically prompted by $P_\text{critic1}$ to give feedback $R_1$ on the numbers extracted to answer the question.

Following this, the context, along with the feedback $R_1$ from the first critic agent, is passed to the second critic agent, $\text{agent}_\text{critic2}$. This agent, guided by its prompt $P_\text{critic2}$, reviews the calculation steps and reasoning process used to derive the final answer, producing feedback $R_2$.

Lastly, both refinement messages $R_1$ and $R_2$ are passed to the financial expert agent, $\text{agent}_\text{expert}$, which then refines its answer based on this feedback, producing $A_\text{revised}$. This multi-layered critique process allows for a more comprehensive review of both the data extraction and the reasoning steps, potentially leading to more accurate responses.


\subsection{Implementation Details}
All multi-agent systems in our study are implemented using the AutoGen framework \cite{wu2023autogen}, an open-source system for multi-agent development. The executor agent is constructed using the UserProxyAgent class from AutoGen, while the financial expert and critic agents are implemented using the AssistantAgent class. Our study focuses on improving the numerical reasoning capabilities of open-source LLMs through collaborative agent interactions.
In line with this objective, we equip the financial expert and critic agents with LLMs and experiment with two settings. In the first setting, we use LLaMA3-8B for both the financial expert and critic agents. To investigate the effect of model size, we conduct a second set of experiments using the larger LLaMA3-70B model. For all experiments, the temperature parameter is set to 0.1 to ensure consistent outputs from the models.
This experimental setup allows us to evaluate the impact of refinement and specialized critique on the quality and accuracy of financial analysis and question-answering tasks.
\section{Experiments}
To validate the effectiveness of our proposed multi-agent framework, we conduct comprehensive experiments comparing it against several state-of-the-art models.

\subsection{Datasets and Evaluation Metrics}
We apply the proposed framework to three popular tabular and textual QA datasets that require numerical reasoning. These datasets are:

\begin{itemize}
\item \textbf{FinQA} \cite{chen2021finqa}: An expert-annotated dataset for financial QA, featuring tabular and textual data from financial reports. It tests numerical reasoning skills including addition, subtraction, multiplication, division, and numerical comparison. We use the test set, which comprises 1,147 questions.\footnote{The dataset was obtained from \url{https://huggingface.co/TheFinAI}} 
\item \textbf{ConvFinQA} \cite{chen2022convfinqa}: A dataset derived from FinQA, designed to simulate conversation flow. It employs a framework for decomposing and concatenating of multi-hop questions, creating a more interactive and dialogue-like structure for financial QA tasks. We employ a subset of the dataset comprising 800 questions.\footnote{The dataset was obtained from \url{https://huggingface.co/FinGPT}}
\item \textbf{TAT-QA} \cite{zhu2021tat}: A dataset built from tables and paragraphs extracted from financial reports. While the full test dataset contains 1,669 questions, we focus on the subset requiring numerical reasoning. We use 731 questions (approximately 40\% of the total) that specifically demand numerical reasoning.\footnote{The dataset was obtained from \url{https://huggingface.co/TheFinAI}}
\end{itemize}

The subsets of datasets utilized in this study are accessible through our Hugging Face repository. \footnote{\url{https://huggingface.co/Sorour}} For performance evaluation, we employ the Exact Match (EM) metric. To optimize computational resources, critic agents only reassessed questions incorrectly answered in the single-agent setting. The final performance score for critic agents combines the single agent's correct answers with additional correct answers from the critic agents \cite{renze2024self}.

\subsection{Baseline Methods}
We compare the performance of our multi-agent QA framework, which incorporates smaller LLMs (LLaMA3-8B and LLaMA3-70B), against state-of-the-art fine-tuned LLMs on tabular and textual QA datasets. Our comparison includes:
\begin{itemize}
\item \textbf{TAT-LLM} \cite{zhu2024tat111}: Fine-tuned LLaMA-13B and LLaMA-70B models on a combination of tabular QA datasets. We report the results of the best-performing model.
\item \textbf{HUSKY} \cite{kim2024husky}: An approach that pre-trains an action generator to iteratively predict high-level steps and associated tools for solving tasks across different domains. It then fine-tunes base language models with high-quality training data to integrate highly expert models.
\end{itemize}
Additionally, we contrast our framework with proprietary and larger LLMs, including GPT-4o-mini (one of the fastest and most cost-effective proprietary models), Claude 3.5 Sonnet, and LLaMA3.1-405B in a single-agent setting. This comparative analysis aims to demonstrate the effectiveness of our multi-agent framework, particularly highlighting the impact of the critic agent in improving multi-step reasoning.

 \begin{figure*}[htp]
    \centering
    \includegraphics[width=\textwidth,height=5cm]{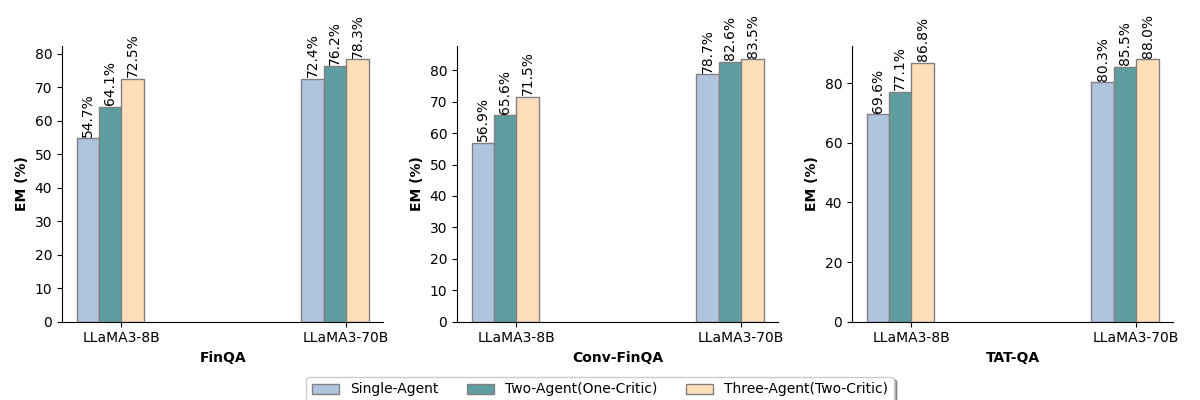}
    \caption{Performance Comparison of Single-Agent, Two-Agent, and Three-Agent Settings Across FinQA, ConvFinQA, and TAT-QA Datasets Using LLaMA3-8B and LLaMA3-70B Models}
    \label{fig6}
\end{figure*}

\begin{table*}
\begin{center}
\begin{tabular}{ l c c c } 
\hline
\textbf{Model} & 
\begin{tabular}{ c } 
\textbf{FinQA} \\
\hline
EM
\end{tabular}
&
\begin{tabular}{ c } 
\textbf{Conv-FinQA} \\
\hline
EM 
\end{tabular}
&
\begin{tabular}{ c } 
\textbf{TAT-QA} \\
\hline
EM 
\end{tabular}
\\
\hline 
\begin{tabular}{ l } 
 LLaMA3.1-405B\\GPT-4o-mini \\ Claude-3.5-Sonnet
\end{tabular}
& 
\begin{tabular}{ c } 
78.29 \\ 74.70 \\ \textbf{84.20}
\end{tabular}
& 
\begin{tabular}{ c } 
85.71 \\ 81.62 \\ \textbf{86.34}
\end{tabular}
& 
\begin{tabular}{ c } 
90.68 \\ 86.73 \\ \textbf{94.21}
\end{tabular}
\\ \hline
\begin{tabular}{ l } 
\textbf{Baseline Models} \\  \hline
TAT-LLM-70B \\ HUSKY-LLaMA
\end{tabular}
&
\begin{tabular}{ c } 
\begin{tabular}{ c } 
\\ 76.81 \\ 20.90 
\end{tabular}
\end{tabular}
& 
\begin{tabular}{ c } 
\\ - \\ - 
\end{tabular}
& 
\begin{tabular}{ c } 
\\ - \\ 42.3 
\end{tabular}
\\ \hline

\begin{tabular}{ l } 
\textbf{LLaMA3-8B} \\ \hline
Single-Agent \\ Two-Agent \\ Three-Agent \\ gain
\end{tabular}
& 
\begin{tabular}{ c } 
\\ 54.67 \\ 64.10 \\ 72.48 \\ \textcolor{teal}{+15.81}
\end{tabular}
& 
\begin{tabular}{ c } 
\\ 56.94 \\ 65.58 \\ 71.49 \\ \textcolor{teal}{+14.55}
\end{tabular}
& 
\begin{tabular}{ c } 
\\ 69.58 \\ 77.12 \\ 86.84 \\ \textcolor{teal}{+17.26}
\end{tabular}
\\ \hline

\begin{tabular}{ l } 
\textbf{LLaMA3-70B} \\ \hline
Single-Agent \\ Two-Agent \\ Three-Agent \\ gain(+)
\end{tabular}
& 
\begin{tabular}{ c } 
\\ 72.36 \\ 76.19 \\ 79.29 \\ \textcolor{teal}{+6.93}
\end{tabular}
& 
\begin{tabular}{ c } 
\\ 78.69 \\ 82.58 \\ 83.45 \\ \textcolor{teal}{+4.76}
\end{tabular}
& 
\begin{tabular}{ c } 
\\ 80.30 \\ 85.49 \\ 87.96 \\ \textcolor{teal}{+7.66}
\end{tabular}
\\ \hline
\end{tabular}
\end{center}
    \caption{Performance comparison of Fine-Tuned TAT-LLM, HUSKY-LLM (results from respective papers \cite{zhu2024tat111, kim2024husky}), and LLaMA models (LLaMA3-8B and LLaMA3-70B) in Single-Agent, Two-Agent, and Three-Agent configurations. Includes multi-agent models (with performance gains over single-agent setup shown in green), as well as single-agent performance of Claude 3.5 Sonnet, LLaMA3.1-405B, and GPT-4-mini. Our results span across the FinQA, ConvFinQA, and TAT-QA datasets.}
\label{table1}
\end{table*}

\section{Main Results}
Table \ref{table1} and Figure \ref{fig6} summarize experimental outcomes. Our analysis reveals several key findings:

\textbf{Single-Agent Performance:}
The single-agent system utilizing LLaMA3-8B substantially underperformed compared to the TAT-LLM-70B model on the FinQA dataset (Note: For TAT-QA, we employed a subset of the dataset and therefore cannot report comprehensive results). This discrepancy highlights the limitations of smaller LLMs in financial QA tasks requiring numerical reasoning. Our examination of the responses indicated that the model often lacked sufficient financial knowledge to accurately interpret questions. In some instances, while it successfully extracted correct numerical data from texts and tables, it struggled to execute division operations accurately.

\textbf{Impact of Critic Agent:}
As depicted in Table \ref{table1} and Figure \ref{fig6}, incorporating a critic agent significantly enhanced performance, yielding an average improvement of 10\% across all datasets for the LLaMA3-8B model. This substantial gain underscores the efficacy of our approach in enhancing the capabilities of smaller open-source LLMs in financial QA tasks. Upon examining the critic agent's responses, we observed its ability to identify incorrect steps or miscalculations in deriving answers. However, when the initial question interpretation was inaccurate, the critic agent struggled to provide constructive feedback. The addition of a second critic agent further improved performance, resulting in an average margin increase of 5\% across models.

\textbf{Effect of Model Size:}
In the single-agent configuration, LLaMA3-70B significantly outperformed the 8B version, even surpassing the three-agent setup on the ConvFinQA dataset. This superiority suggests enhanced numerical reasoning capabilities in larger models. Despite this improvement, LLaMA3-70B still encountered challenges in comprehending questions and accurately extracting numerical data from tables, though these issues were less prevalent compared to LLaMA3-8B. The larger model also exhibited fewer calculation errors.

The introduction of a critic agent to the LLaMA3-70B model yielded more modest improvements of 3.83\%, 3.89\%, and 5.19\% for the FinQA, ConvFinQA, and TAT-QA datasets, respectively. This smaller enhancement can be attributed to the model's inherently lower rate of calculation errors, which limits the scope for improvement through error correction.

\textbf{Comparison with Larger LLMs:}
Among the single-agent performances of LLaMA3.1-405B, GPT-4o-mini, and Claude 3.5 Sonnet, the Sonnet model significantly outperformed its counterparts, showcasing impressive logical and numerical reasoning abilities in financial question comprehension and answer derivation. Notably, the LLaMA3.1-405B model marginally outperformed the GPT-4o-mini model, possibly due to more comprehensive training data that may have included financial information.
Remarkably, our framework, particularly the three-agent LLaMA3-70B setting, performed comparably to or even surpassed larger and proprietary LLMs. This outcome highlights the effectiveness of our refinement process in enhancing smaller LLMs' reasoning and calculation capabilities for financial QA tasks.

\textbf{Performance Across Datasets:}
We observed superior performance on ConvFinQA, especially for LLaMA3-70B, LLaMA3.1-405B, GPT-4o-mini, and Claude 3.5 Sonnet models. This dataset presents each intermediate step as a separate question, with answers provided for multi-step problems. This format minimizes the risk of incorrect information extraction from texts or tables, leading to higher overall performance compared to the FinQA dataset. A similar trend was noted for TAT-QA, which includes simpler questions, some requiring extraction of only a single number from a table without complex calculations.

\textbf{Conclusion:}
Our results demonstrate the efficacy of the proposed framework in significantly enhancing the performance of smaller LLMs. The framework's output surpassed state-of-the-art fine-tuned LLMs, illustrating how our LLM-based agent approach can rival extensively trained models without the need for substantial computational resources. Furthermore, the results were comparable to, though slightly below, powerful LLMs like Claude 3.5 Sonnet. This suggests that our method could potentially reduce reliance on costly API calls to these advanced proprietary models while maintaining competitive performance in financial QA tasks.

\section{Conclusions and Future Work}
In this study, we introduced a multi-agent framework for financial QA tasks, demonstrating its effectiveness in enhancing the numerical reasoning capabilities of smaller LLMs over tabular and textual data. Our approach, which incorporates a critic agent for refinement, significantly improved the accuracy of smaller LLMs. Notably, our method achieved performance comparable to state-of-the-art fine-tuned models without the need for extensive computational resources for training. Although it slightly underperformed compared to the most advanced models like Claude 3.5 Sonnet, our framework offers a cost-effective alternative for tackling complex financial QA tasks, potentially reducing reliance on costly API calls to advanced LLMs.
In future work, we aim to develop a multi-step expert agent for extracting and performing calculation steps, allowing us to observe the impact of multiple agents in deriving answers. Additionally, we plan to experiment with an iterative (multi-turn) refinement process, similar to a debate-like discussion, to examine the performance of a debate framework as explored in previous studies \cite{lin2024interpreting}.


\bibliographystyle{ACM-Reference-Format}
\bibliography{sample}

\appendix
\section{Appendix}
\label{appendix}
The image demonstrates the interaction between a financial expert agent and a critic agent. The workflow begins with the financial expert agent providing an initial answer to a given financial question. The critic agent then reviews this response, offering detailed feedback on the accuracy of calculations and the interpretation of financial data. Finally, the financial expert agent uses this critique to refine its answer, correcting any identified errors and improving the overall response. This iterative process showcases how the multi-agent framework enhances the accuracy of financial QA tasks, particularly in handling complex numerical reasoning and data interpretation challenges.
\begin{figure*}[htp]
    \centering
    \includegraphics[width=\textwidth]{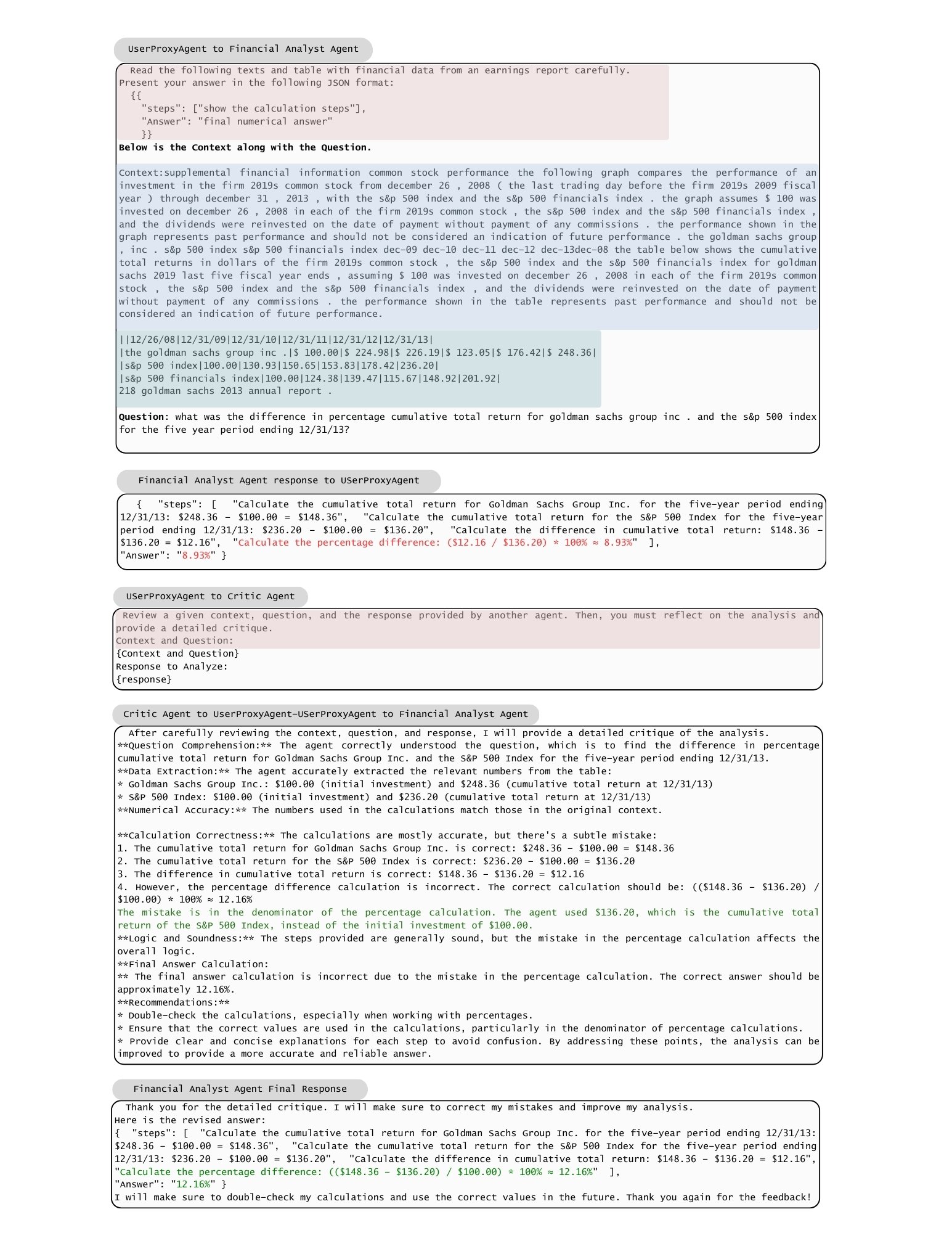}
    \caption{Workflow of the Two-Agent Setting with Refinement Process }
    \label{fig7}
\end{figure*}






\end{document}